\documentclass[final]{cvpr}

\usepackage{times}
\usepackage{epsfig}
\usepackage{graphicx}
\usepackage{amsmath}
\usepackage{amssymb}
\usepackage{multirow}
\usepackage{tabularx}
\usepackage{bm}
\usepackage{hyphenat}
\usepackage[ruled,linesnumbered]{algorithm2e}
\usepackage{array}
\makeatletter
\newcommand{\thickhline}{%
	\noalign {\ifnum 0=`}\fi \hrule height 0.1em
	\futurelet \reserved@a \@xhline
}
\makeatother
\usepackage[pagebackref=true,breaklinks=true,colorlinks,bookmarks=false]{hyperref}

\def\cvprPaperID{3292} % *** Enter the CVPR Paper ID here
\def\confYear{CVPR 2021}
\def\etal{\textit{et al.}}
\def\ie{\textit{i.e.}}
\def\eg{\textit{e.g.}}
\def\etc{\textit{etc.}}
\def\X{\mathcal{X}}
\def\Y{\mathcal{Y}}
\def\F{\mathcal{F}}
\def\M{\mathcal{M}}
\def\P{P}
\makeglossary

\begin{document}
\hyphenpenalty=5000
\tolerance=1000
%%%%%%%%% TITLE
\title{Multi-Source Domain Adaptation with Collaborative Learning for Semantic Segmentation}

\author{Jianzhong He\textsuperscript{1,2}\quad
Xu Jia\textsuperscript{3}\footnotemark[1]\quad
Shuaijun Chen\textsuperscript{2}\quad
Jianzhuang Liu\textsuperscript{2}\\
\textsuperscript{1}Data Storage and Intelligent Vision Technical Research Dept, Huawei Cloud.\\
\textsuperscript{2}Noah's Ark Lab, Huawei Technologies.
\textsuperscript{3}Dalian University of Technology.\\
{\tt\small \{jianzhong.he, chenshuaijun, liu.jianzhuang\}@huawei.com, xjia@dlut.edu.cn}\\
%{\textsuperscript{1}Data Storage and Intelligent Vision Technical Research Dept, Huawei Cloud \textsuperscript{2}Noah's Ark Lab, Huawei Technologies}
}

\maketitle

\renewcommand{\thefootnote}{\fnsymbol{footnote}}
\footnotetext[1]{Corresponding author}

%%%%%%%%% ABSTRACT
\begin{abstract}
	Multi-source unsupervised domain adaptation~(MSDA) aims at adapting models trained on multiple labeled source domains to an unlabeled target domain. In this paper, we propose a novel multi-source domain adaptation framework based on collaborative learning for semantic segmentation. Firstly, a simple image translation method is introduced to align the pixel value distribution to reduce the gap between source domains and target domain to some extent. Then, to fully exploit the essential semantic information across source domains, we propose a collaborative learning method for domain adaptation without seeing any data from target domain. In addition, similar to the setting of unsupervised domain adaptation, unlabeled target domain data is leveraged to further improve the performance of domain adaptation. This is achieved by additionally constraining the outputs of multiple adaptation models with pseudo labels online generated by an ensembled model. Extensive experiments and ablation studies are conducted on the widely-used domain adaptation benchmark datasets in semantic segmentation. Our proposed method achieves 59.0\% mIoU on the validation set of Cityscapes by training on the labeled Synscapes and GTA5 datasets and unlabeled training set of Cityscapes. It significantly outperforms all previous state-of-the-arts single-source and multi-source unsupervised domain adaptation methods.
   %To fully  exploit the common knowledge of source domains, we propose a Collaborative learning method which each single source trained model learn from the other models. Thus the each model not only need to correctly learn the source domain of itself but also the others, which enforce the model extract common information for segmentation. Furthermore, to utilizing the unlabeled target data, we propose a mutual learning approach, which help models to learn feature of target. Extensive experiments and ablation studies are conducted on the widely-used domain adaptation settings in semantic segmentation. Our proposed method achieves 58.8\% mIoU on the validation set of Cityscapes by training on the Synscapes and GTA5 datasets with unlabeled training set of Cityscapes, and significantly outperforms all the previous state-of-the-arts single-source and multi-source unsuoervised domain adaptation methods.
\end{abstract}

%%%%%%%%% BODY TEXT
\section{Introduction}

\begin{figure}[t]
	\begin{center}
		\includegraphics[width=1.02\linewidth]{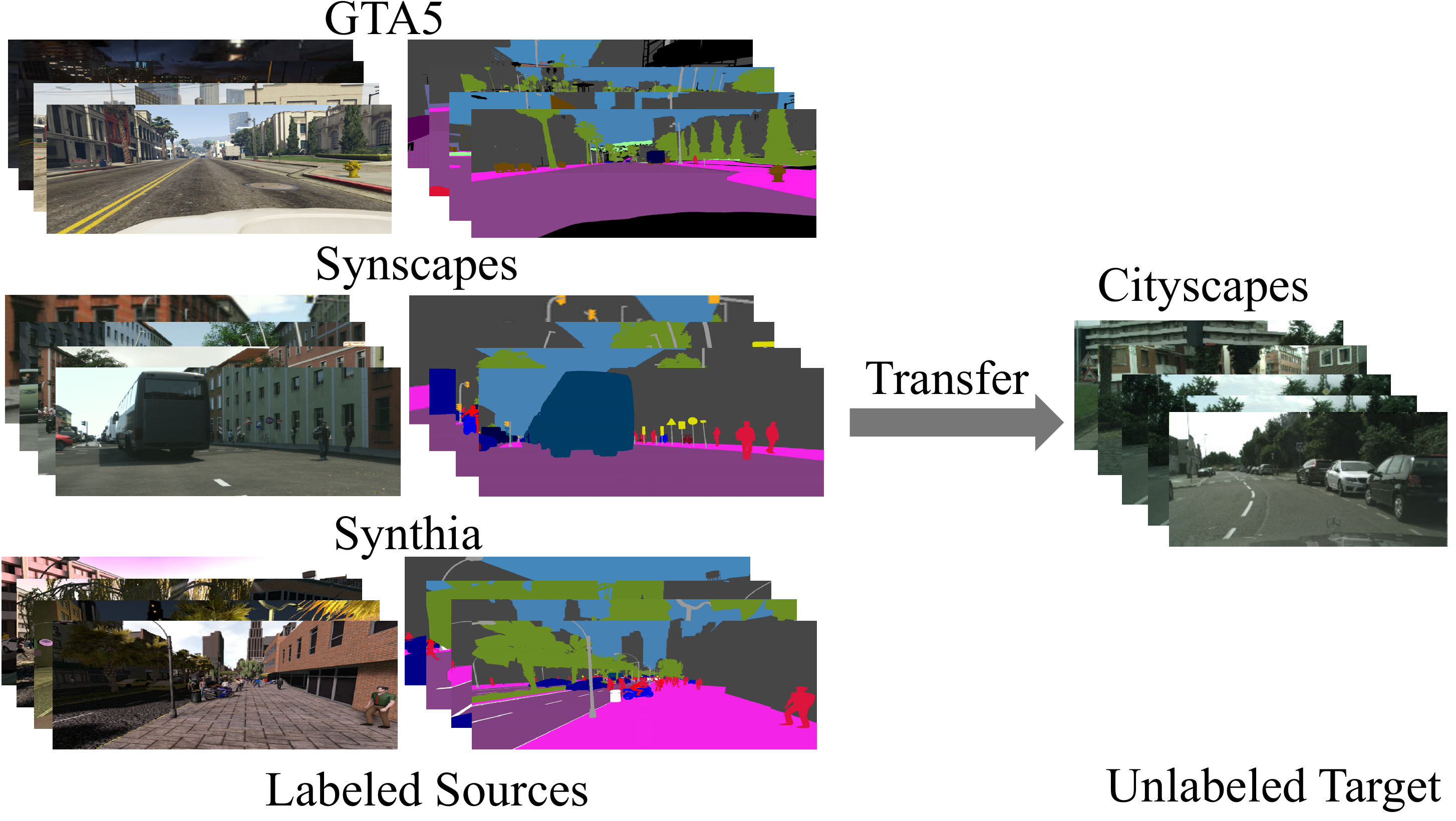}
	\end{center}
	%\caption{The image comparisons of different synthetic sources, transformed images and the real target. The first column is the origin images of GTA5 or Synscapes and the last column correspond to Cityscapes. The middle two columns are the transformed images that based on RGB and LAB color space respectively.The region in color boxes contains numerical artifacts after image translation.}
	\vspace{-6pt}
	\caption{Multi-source domain adaptation for semantic segmentation. The left shows synthetic images and corresponding labels generated from different simulators, which suffer domain shift between each other but share similar semantic contexts. The right part shows unlabeled target images sampled from real scenes.}
	\label{fig:images_comp}
	\vspace{-10pt}
\end{figure}

Semantic segmentation as one of the core tasks in computer vision community, aims to  assign semantic label to each pixel of images, \textit{e.g.}, person, car, road and \textit{etc.}. With the development of convolutional neural networks~(CNNs), semantic segmentation has made great progress recently. For example, recent deep methods~\cite{chen2018encoder,fu2019dual,he2020bdcn,Li_2020_CVPR,zhao2017pyramid}, have achieved superior performance on almost all public benchmarks. However, their success is based on the large numbers of densely annotated images which used to train the networks. Dense pixel-level annotation for semantic segmentation is very laborious and expensive, \eg, annotating one image in the Cityscapes dataset~\cite{cordts2016cityscapes} takes about 90 minutes, which makes it difficult and sometimes even impossible to collect large amounts of densely annotated images for semantic segmentation. Thanks to the recent progress in graphics and simulation infrastructure, simulators can generate lots of images with dense annotation for semantic segmentation, such as recent proposed large-scale dense labeled datasets SYNTHIA~\cite{Ros_2016_CVPR}, GTA5~\cite{Richter_2016_ECCV} and Synscapes~\cite{wrenninge2018synscapes}. Although the huge amounts of annotated synthesized images are very close to the real scene, there is still great domain gap between synthetic datasets and real scene datasets. The domain gap causes another problem that networks trained on synthetic datasets often perform poorly on real target scenes. To handle this issue, many un-/semi-supervised domain adaptation~(UDA) approaches are proposed, like ~\cite{li2019bidirectional,pan2020unsupervised,chen2021dual, takashi2021mtda,tsai2018learning,vu2019advent,FDA_Yang_2020_CVPR} and \textit{etc.}, with the purpose of mitigating the gap between synthetic source and real target domain. Over the past years, UDA has made a great progress.

Although existing works have greatly boosted the performance of UDA for semantic segmentation, most of them focus on single source. Seldom works consider a more practical setting where labeled datasets from multiple sources with different distributions are available, \eg, SYNTHIA and GTA5. Training with multiple sources can further alleviate the problem on lack of annotated data. Moreover, multiple sources sampled from different distribution can also encourage networks to learn more essential knowledge for semantic segmentation. A straightforward approach is to simply combine all source domains into a single one, and then trains a UDA model on the combined sources and target domain dataset. This simple method can indeed boost the performance, but it does not fully exploit the abundant information across multiple source domains. Domain shift across multiple sources restricts the power within them in learning a more powerful domain adaptation model.

There are several multi-source deep UDA methods are proposed recently to exploit multiple source domains for better adaptation. 
They align different domains by translating images from source domains to the target style via generative adversarial networks~(GAN). However, most of them~\cite{Peng_2019_ICCV, sun2011two, liu2016structure} work on image classification task except for MADAN~\cite{zhao2019multi} which works on semantic segmentation, a pixel-wise prediction task.
In this paper, we propose an approach based on collaborative learning and image translation to address multi-source domain adaptation for semantic segmentation. 

Our observation shows that appearance discrepancy especially color discrepancy between source domains and target domain has a great impact on the performance of adaptation. Existing works~\cite{ulyanov2017improved,li2018closed,yoo2019photorealistic,huang2017arbitrary} demonstrate that style transfer could reduce this discrepancy in some extent. However, most of them are complicated to plug in networks during training process. Therefore, we propose a simple image translation method to first mitigate domain gap between sources and target. Unlike MADAN~\cite{zhao2019multi}, FDA~\cite{FDA_Yang_2020_CVPR} and GAN-based translation methods, we propose to translate source domain images to the target style by aligning different distributions to the target domain in LAB color space. In addition, we observe that apart from discrepancy in appearance, images from different domains do still share much similarity in semantic contexts as shown in Fig~\ref{fig:images_comp}. The shape of instances~(person, car, bike and \etc) and spatial layout of different instances~(cars always on the road, sidewalk adjacent to the road, sky on the top and \etc) are almost the same in all domains. Two collaborative learning strategies are proposed to explore essential and domain-invariant semantic contexts across different domains. 
%First, we investigate the case of domain adaptation without seeing any data from target domain, 
First we propose a collaborative learning between source domains to investigate the case of domain adaptation without seeing any data from target domain,
which is also called domain generalization in previous works~\cite{dlow_gong2019dlow,pan2018two,zhang2020generalizable}. 
For each source domain, we have a semantic segmentation network supervised by annotation maps, and an additional soft supervision coming from other models trained on a different source domain. 
In addition, similar to previous UDA methods~\cite{zhao2019multi,FDA_Yang_2020_CVPR}, we also consider making full use of the unlabeled data of the target domain to further boost the performance. A collaborative learning based on target domain is proposed, in which an ensemble of models trained on source domains is used to produce pseudo labels for data from target domain in an online fashion. In turn, each model can be additionally supervised by the generated pseudo labels. Such two collaborations help constantly improve each model's adaptation capability to target domain during the training process.

The performance of our method significantly outperforms other state-of-the-art single-source and multi-source UDA methods. This success of proposed method is mainly attributed to the effective image translation and domain-invariant feature learning. Note that, our method can be trained in both end-to-end and stage-wise.

%------------------------------------------------------------------------
\section{Related works}
In this section, we briefly review some related works in the literature, \textit{i.e.}, semantic segmentation, domain generalization and unsupervised domain adaptation.

\subsection{Semantic Segmentation}
Semantic segmentation plays a vital role in computer vision community and is beneficial to many practical applications, such as autonomous driving, virtual reality and medical imaging and \etc. It has developed several decade years and is well researched. Since Long \etal~\cite{long2015fully} propose to transform the classification CNNs to fully convolutional network for semantic segmentation, large numbers of deep learning based methods have been proposed and greatly boost the advances of this task. For example, Chen \etal propose the DeepLab series~\cite{chen2017deeplab,chen2018encoder} approaches which utilize \'atrous spatial pyramid pooling~(ASPP) to capture different scale of context information. Fu \etal~\cite{fu2019dual} involve non-local attention block into the CNNs architecture to exploit the global context of image and relation of objects. Hou \etal~\cite{Hou_2020_CVPR} propose a new effective strip pooling to model long-range dependencies. % and collect rich global contextual information. 
However, the advanced performance of these semantic segmentation methods often build on the large amounts of densely annotated images which are usually hard and sometimes impossible to collect. 

\begin{figure*}[t]
	\begin{center}
		\includegraphics[width=0.88\linewidth]{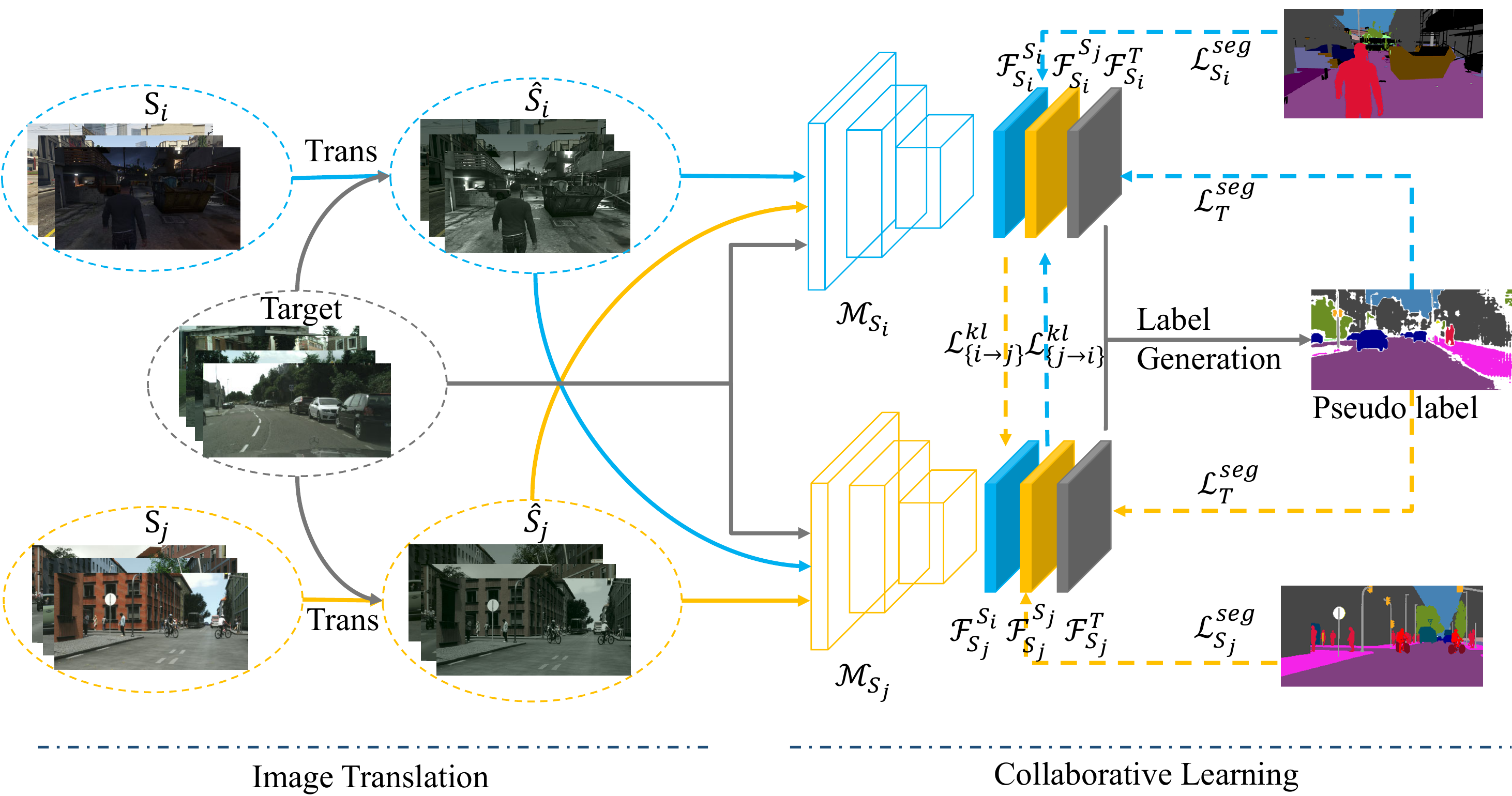}
	\end{center}
\vspace{-6pt}
	%\caption{The overall architecture of proposed approach. The solid arrows represent the forward data flow and different colors indicate different sources or target flow. The dash arrows represent gradient backward of flow and different colors indicate flows for different models. For simplicity, here we just show the case of two sources and not show the process of source image transformation to target.}
	\caption{The overall framework of proposed approach consists of three components, including that image-to-image translation based on LAB color space, collaborative learning between source domains and collaborative learning on target domain. The solid arrows represent the forward data flow and different colors indicate different source domains or target domain data flow. The dash arrows represent the supervision to the network outputs. For illustration, we just show the case of two source domains as an example to explain our method.}
	\label{fig:overall_arch}
	\vspace{-10pt}
\end{figure*}

\subsection{Domain Generalization}
Domain generalization is a particular case of transfer learning. It's purpose is to enhance the generalization ability of models on new domains that have not been seen during the training process. Currently, most of the domain generalization methods can be categorized to three parts: data-based, feature-based and meta-learning based. Most data-based methods employ GANs, Variational Autoencoder~(VAE) or other image edit methods to generate new data for network training to enhance robustness and generalization of the models, \eg, Dlow~\cite{dlow_gong2019dlow} and STRG~\cite{yue2019domain}. Then, the feature-based methods mainly aim to learn representations invariant to different domains by adversarial learning to align features or employing normalization to eliminate the style information, like CADAG~\cite{rahman2020correlation} and IBN-Net~\cite{pan2018two}. Meta-learning based methods aim to enhance the generalization ability by using meta learning, such as Zhang \etal~\cite{zhang2020generalizable}  deal with the domain generalization from the training scheme perspective and develop a target-specific normalization method to further boost the generalization ability in unseen target domain. Domain generalization do not access the data in target domain which may not learn the optimal feature for target domain.

\subsection{Unsupervised Domain Adaptation}
In contrast to domain generalization, unsupervised domain adaptation involves in unlabeled data of target domain during training process to learn knowledge of it. UDA has developed many years and large numbers of methods have been proposed. From the perspective of the number of source domains, domain adaptation methods can be split into two categories: single-source and multi-source.

\textit{Single-source domain adaptation} focuses on single-source-single-target setting. Most of existing UDA works are for classification, like MMD~\cite{long2015learning,long2016unsupervised}, ADDA~\cite{tzeng2017adversarial} and \etal. With the synthetic dataset GTA5, SYNTHIA and Synscapes proposed , UDA for semantic segmentation has also achieved great advances in recent. For example, Tsai~\etal~\cite{tsai2018learning} propose AdaptSeg that based on adversarial learning to aligns scene layout and local context of images between source and target domain. 
%Vu \etal~\cite{vu2019advent} propose to minimize the entropy loss of images outputs in target domain and further boost the performance of UDA for semantic segmentation. 
Yang \etal~\cite{FDA_Yang_2020_CVPR} draw on the image-to-image translation and propose to using Fourier Transform to translate images in source domain to the style of target domain. While Li \etal~\cite{li2019bidirectional} propose a bidirectional framework for domain adaptation and learn a image translation model with perceptual loss to translate images to target style. These methods have significantly boost the performance of UDA in semantic segmentation. However, they do not take the existing multiple different source domains into account which is a great waste of labeled resources.
%As shown in~\cite{zhao2019multi}, simply combine them to one and adopt the single source UDA methods may not work well.

\textit{Multi-source domain adaptation} aims to make full use of existing labeled source for adaptation. Compared to single-source UDA, it is relatively more challenging because of the shift between source domains. There are also some multi-source UDA works but most of them focus on image classification. Directly extending these methods for semantic segmentation may not work. Zhao \etal~\cite{zhao2019multi} propose MADAN for semantic segmentation. They first learns a cycle GAN~\cite{zhu2017unpaired} for each source domain to translate images to target, and then combine different adapted sources with specific weights. Finally, training a adaptation network on the combined data in a way similar to single-source UDA. In contrast to~\cite{zhao2019multi}, we not only draw on the style-transfer for image-to-image translation, also aim to learn the shared semantic information of source domains and explore the knowledge of unlabeled target data.

\begin{figure}[t]
	\begin{center}
		\includegraphics[width=1\linewidth]{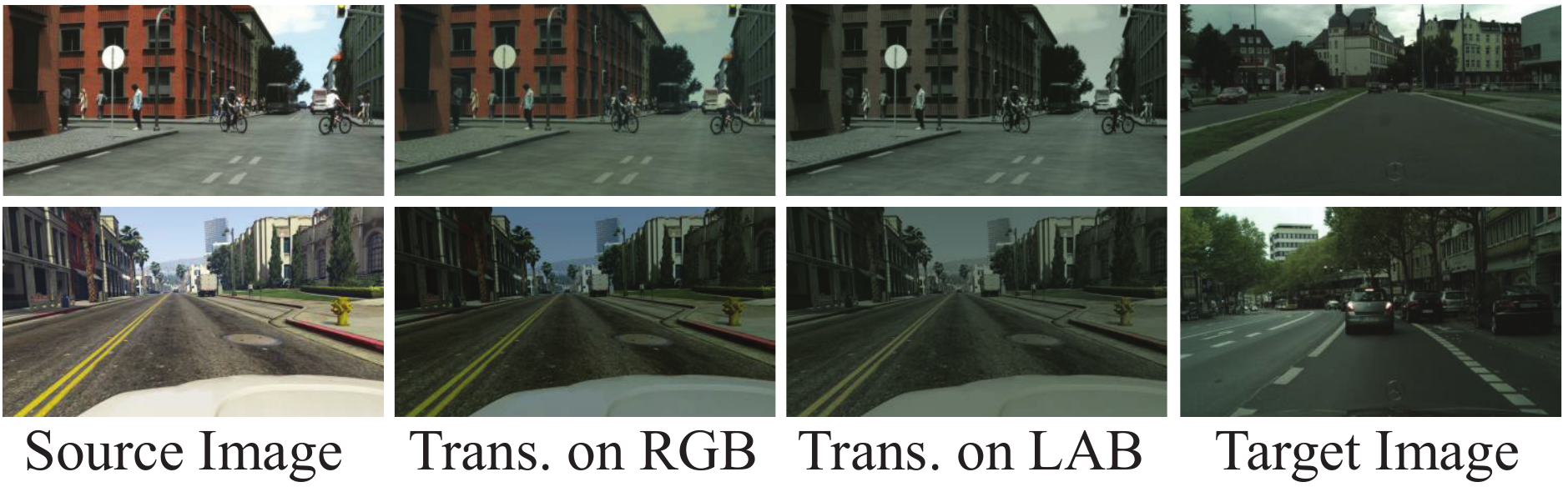}
	\end{center}
	\vspace{-4mm}
	\caption{The qualitative comparison of image translation on different color space.% Each column corresponds to origin source images, translated based on RGB and LAB color space and the selected target images.
	}
	\label{fig:trans_comp}
	\vspace{-5mm}
\end{figure}
\section{Proposed Approaches}
%The overall architecture of proposed approach are shown in Fig.~\ref{fig:overall_arch}. In summary, our method consist of three parts, \ie, image transformation based on LAB color space, Collaborative learning on the sources and mutual learning on the target domain. Our method can be trained with both end-to-end and stage-wise strategies.
%\subsection{Formulation}
\subsection{LAB-based Image Translation}
\label{sec:lab}
%\noindent\textbf{LAB-based image transformation.} 
We observe that the domain discrepancy between domains mainly lies in appearance of images, \ie, color and texture. For example, the appearance discrepancy between source domains~(GTA5 and SYNTHIA), and between source domains and target domain~(GTA5 and Cityscapes) as demonstrated in Fig.~\ref{fig:images_comp}. This discrepancy would further increase the difficulty of multi-source domain adaptation. An effective style transfer could reduce this discrepancy to some extent. For simplicity and efficiency, we proposed an image translation method that translates the style of images in source domains to the style of target domain by aligning the distribution of pixel values, like~\cite{reinhard2001color}. Specifically, due to the gamut of LAB color space is larger than RGB color space, and the style of images translated on LAB color space are closer to the style of images in target domain than directly operated on RGB color space as shown in Fig.~\ref{fig:trans_comp}, we involve the image translation based on LAB color space to achieve the target of reducing domain discrepancy. 
%We translate images of source domains to the style of target domain based on LAB color space, which makes the translated images similar to the target in appearance, as shown in Fig~\ref{fig:overall_arch}.

Specifically, for a RGB image $\X_S^{RGB}$ in source domains, it is firstly converted to LAB color space to generate LAB image $\X_S^{LAB}$, $\X_S^{LAB} = rgb2lab(\X_S^{RGB})$. Then, we calculate the mean~$\mu_S$ and standard deviation values~$\sigma_S$ of each channel of the generated LAB image $\X_S^{LAB}$. At the same time, an image from target domain is randomly selected and converted to LAB color space in same way as image from source. After that, we also calculate the mean $\mu_T$ and standard deviation values $\sigma_T$ of this converted target image. Finally, we translate the converted LAB image from source domains to the style of target by shifting the distribution of pixel values to the image of target domain, \textit{ie.},

\begin{equation}
	\hat{\X}^{LAB}_S = \frac{(\X_S^{LAB} - \mu_S)}{\sigma_S} * \sigma_T + \mu_T.
\end{equation}
After aligning distribution in LAB color space, we then convert the translated LAB image~$\hat{\X}^{LAB}_S$  from source domains back to the RGB color space $\hat{\X}_S^{RGB}$, $\hat{\X}_S^{RGB}=lab2rgb(\hat{\X}_S^{LAB})$, for the subsequent training. The $rgb2lab$ and $lab2rgb$ are the library conversion function in python, and also easy to implement it with CUDA to accelerate computation. Fig~\ref{fig:trans_comp} shows the qualitative comparison between RGB- and LAB-based translation. We can see that the images after translation are closer to the style of target domain, which the gap between domains is reduced to some extent. Moreover, the LAB-based translated images are closer to the style of target images.
%Fig.~\ref{fig:images_comp} shows the comparison of transformed images that based on RGB and LAB, which the LAB transformed images are closer to target and are not be broken.

%\vspace{2mm}
%\noindent\textbf{Collaborative Learning.}
\subsection{Collaborative Learning between Labeled Source Domains}

As shown in Fig.~\ref{fig:images_comp}, although the images of different source domains are sampled from different simulators, \ie, from different \textit{i.i.d} distributions, they are highly structured and share many similarities, \eg, the spatial layout of different categories and local context of objects. Based on this observation and to encourage models learn these essential similar properties for semantic segmentation, we propose a collaborative learning~(Co-Learning) method which fully takes advantage of the capacity of different models learned on different source domains. This approach utilizes a model learn from one source domain to teach the model trained on another one, which allows that model to learn the knowledge from current source domain. 

Assume that there are $N$ different labeled source domains $S=\{S_1, S_2, \cdots, S_N\}$ which are sampled from $N$ different \textit{i.i.d} distributions, and $N$ deep neural networks $\M=\{\M_{S_1}, \M_{S_2}, \cdots, \M_{S_N}\}$ of the same architecture but different weights learned on these source domains. Then, for an model $\M_{S_i}$, the learning process of model $\M_{S_i}$ is supervised by segmentation loss on labeled data from source $S_i$ and collaborative loss on output from source $S_{k, k\ne i}$. That is, for model $\mathcal{\M}_{S_i}$, the object function is 
\begin{equation}
%\small
%	\begin{split}
%		\mathcal{L}_i &= \mathcal{L}_{CE}^{S_i} + \lambda_{col}~\mathcal{L}_{kl}^{i\rightarrow j} ~, \\
%		\mathcal{L}_j &= \mathcal{L}_{CE}^{S_j} + \lambda_{col}~\mathcal{L}_{kl}^{j\rightarrow i}
%	\end{split}
%\mathcal{L}_i = \mathcal{L}_{ce}^{S_i}(\F_i^{S_i}, \Y^{S_i}) + \lambda_{col}~\sum_{k,k\neq i}\mathcal{L}_{kl}^{k\rightarrow i}(\F_i^{S_k}\|\F_k^{S_k}),
\mathcal{L}_{i} = \mathcal{L}^{seg}_{S_i}(\F_{S_i}^{S_i}, \Y_{S_i}) + \lambda_S^{col}\mathcal{L}_S^{col}(\{(\F_{S_i}^{S_k}, \F_{S_k}^{S_k})_{k\ne i}\}),
\end{equation}
where the loss $\mathcal{L}^{seg}$ is the cross entropy loss, \ie,
\begin{equation}
	%\small
	\label{equ:ce}
	\mathcal{L}^{seg}_{S}(\F_S, \Y_S) = -\frac{1}{|\X_S|}\sum_{h,w}\sum_{c\in C} \Y^{(h,w,c)}_{S}log(\sigma(\F^{(h,w,c)}_{S})),
\end{equation}
and the loss $\mathcal{L}^{col}$ is the average of Kullback-Leibler~(KL) divergence loss, \ie,
\begin{equation}
	%\small
	\label{equ:col}
	\mathcal{L}_S^{col}(\{(\F_{S_i}^{S_k}, \F_{S_k}^{S_k})_{k\ne i}\}) = \frac{1}{N-1}\sum_{k,k\neq i}\mathcal{L}^{kl}_{k\rightarrow i}(\F_{S_k}^{S_k}\|\F_{S_i}^{S_k}),
\end{equation}
\begin{equation}
	%\small
	\label{equ:kl}
	\mathcal{L}^{kl}_{k\rightarrow i}(\F_{S_k}^{S_k}\|\F_{S_i}^{S_k}) = -\frac{1}{|\X_{S_k}|}\sum\sigma(\F_{S_k}^{S_k})\log(\frac{\sigma(\F_{S_i}^{S_k})}{\sigma(\F_{S_k}^{S_k})}).
\end{equation}
$\X, \F, \Y$ are the input image, outputs of networks and the corresponding ground truth respectively. For $\F_{S_i}^{S_k}$, the subscript indicates that $\F$ is generated by $\M_{S_i}$ while the superscript indicates $\F$ is the feature computed for images from domain $S_k$, \ie, $\F_{S_i}^{S_k} = \M_{S_i}(\X_{S_k})$. $C$ is the number of categories to be segmented, $\sigma(\cdot)$ indicates the softmax function, $|\X|$ represents the number of pixels in image $\X$. Collaborative learning can allow the model $\M_{S_i}$ learn the knowledge of other models $\M_{S_{k, k\neq i}}$ learned from corresponding source $S_k$. Thus, rather than stuck in the source domain $S_i$, model $\M_{S_i}$ also tries to learn essential properties between all the source domains for better generalization and segmentation on target domain.
%For simplicity and clear, we show the case of two sources and it is easy to extend to more sources.

\textbf{Difference with existing works:} In the co-training of CLAN~\cite{luo2019taking} and CT~\cite{qin2019generatively}, two diversified classifiers are produced to make predictions and their ensemble with either summation or multiplication is taken as the final prediction; however, in the collaborative learning framework, two segmentation models trained based on each source domain \textbf{teach each other} to extract essential semantic information across domains.
CrDoCo~\cite{chen2019crdoco} uses bi-directional KL divergence to encourage segmentation consistency between images in unlabeled target domain and their source-style transformed correspondences.  
However, in our work, images from both source domains are first transformed to the same style, i.e., target domain style. Then an arbitrary image is sent to two models that are trained for source 1 and source 2 respectively.  
The prediction of the model trained for the source domain that image comes from works as teacher to supervise the other model's prediction on the \textbf{same} image.

%\vspace{2mm}
%\noindent\textbf{Mutual Learning.}
\subsection{Collaborative Learning on Unlabeled Target Domain}
In practice, unlabeled data is often relatively easy and cheap to collect. Moreover, models trained on the data of target domain can learn better features and perform well on target domain. Therefore, we propose a collaborative learning method to fully take advantage of the unlabeled images in target domain, and further boost the performance of models on target domain. 

Denote target domain as $T$ which has collected large amounts of unlabeled images $\{\X_{T}^j\}_{j=1}^{N_T}$. For an image $\X_T$ in the target domain $T$, we first feed it to all the $N$ models $\M=\{\M_{S_i},i=1,2,\cdots,N\}$ and compute the corresponding outputs of models $\M_{S_i}$ for the input image $\X_T$ as $\F_{S_i}^T$, \ie, $\F_{S_i}^T = \M_{S_i}(\X_T)$. Outputs of these networks are ensembled and a softmax function $\sigma(\cdot) $ is used to compute the probability map, \ie,
\begin{equation}
	\hat{\P} = \sigma(\frac{1}{N}\sum_i \F_{S_i}^T).
	\vspace{-3mm}
\end{equation}
Finally, we generate one-hot pseudo labels $\hat{\Y}$ for the target image $\X_T$ by utilizing the probability map in a way similar to ~\cite{li2019bidirectional,zou2018unsupervised}, and the detail process is described in the Algorithm~\ref{algo:gen_labels}. After then, we can train models ${\M_{S_i}}$ with both source domains and target domain. Thus, the object function for model $\M_{S_i}$ becomes
\begin{equation}
	\mathcal{L}_{i} = \mathcal{L}^{seg}_{S_i}(\F_{S_i}^{S_i}, \Y_{S_i}) + \frac{cur\_it}{max\_its}\lambda_T^{seg}\mathcal{L}^{seg}_{T}(\F_{S_i}^{T}, \hat{\Y}_T),
\end{equation}
where cross entropy loss is employed to compute the loss $\mathcal{L}_T^{seg}$ and is same as Equ.~\ref{equ:ce}, the only difference is that the target labels are generated pseudo labels.
Here the the weight term $\frac{cur\_it}{max\_its}$ is designed to prevent poor predictions of the target image in early training process collapsing the network training, $cur\_it$ and $max\_its$ represent current and maximum iterations of training process.
%Note that, the model $M_i$ is same as refereed in the previous section that corresponding training on source $S_i$.

\begin{algorithm}[t]
	\caption{Pseudo Labels Generation}
	\label{algo:gen_labels}
	\KwData{The probability map $\hat{\P}\in \mathcal{R}^{C\times H \times W}$, keep proportion $\alpha$, maximum thresh $\tau$, the ignore label $l_{ig}$}
	\KwResult{one-hot hard pseudo labels $\hat{\Y}$}
	$\hat{\Y}$ $\leftarrow$ $argmax$($\hat{\P}, dim=0$), $\hat{\Y}\in \mathcal{R}^{H\times W}$\\
	\For{c $\leftarrow$ 0 to $C-1$}{
		$\hat{\P}_c \leftarrow sort(\hat{P}_{\{c, \cdot, \cdot\}}, order=Descending)$;\\
		get the number of pixels $n_c$ which are predicted to category $c$: $n_c \leftarrow sum(\hat{\Y} == c)$;\\
		get the threshold $t$ that used to filter the prediction: $t \leftarrow min(\hat{\P}_c[n_c\times\alpha], \tau)$; \\
		$mask1 \leftarrow \hat{\Y} == c$;\\
		$mask2 \leftarrow \hat{\P}_{\{c, \cdot, \cdot\}} <= t$;\\
		$\hat{\Y}[mask1 \ \&\  mask2] \leftarrow l_{ig}$.
		%Set the points in $\hat{Y}$ which are predicted to $c$ and the corresponding probability letter than $t$ to ignore label.
	}
\vspace{-1.6mm}
\end{algorithm}
%\vspace{-2mm}

\subsection{Network Architecture and Training}
The overall architecture of proposed approach are shown in Fig.~\ref{fig:overall_arch}. In summary, it consists of three parts, \ie, image-to-image translation based on LAB color space, collaborative learning between labeled source domains and on unlabeled target domain. Firstly, images of source domains are translated to the style of target domain as described in Sec.~\ref{sec:lab} during training process. Note that for simplicity, we just showcase the collaborative learning approach for multi-source domain adaptation with two source domains. As shown in Fig.~\ref{fig:overall_arch}, we train a segmentation network, \eg, DeepLab v2~\cite{chen2017deeplab}, for each source domain. These networks can be trained end-to-end which initialized from ImageNet pretrained model, or trained with stage-wise which initialized from source fine-tuned model. For each network, the final overall object function is as follows,
\begin{equation}
	\mathcal{L} = \mathcal{L}^{seg}_{S} + \lambda_S^{col}\mathcal{L}_S^{col} + \frac{cur\_it}{max\_its}\lambda_T^{seg}\mathcal{L}^{seg}_{T}.
\end{equation}
The loss $\mathcal{L}^{seg}_{S}$ ensures the correct functionality of networks for segmentation, while the loss $\mathcal{L}_S^{col}$ allows the network to learn the similarities that shared by source domains. In addition, the loss $\mathcal{L}^{seg}_{T}$ allows the network learn the property of target data. When network training get converged, we have two strategies to obtain the final model. One is to choose the model which achieves the best performance on target domain. The other is to keep all models, and ensemble their outputs when inference. Unless otherwise specified, all refer to the last case.
The following section tests the validity of proposed approach.

\begin{table}[t]
	\centering
	\caption{Validity of the proposed image translation method. The performance comparison with the recent single-source UDA methods trained on images that before and after translation.}
	%\caption{The performance comparison that different single-source UDA methods train with LIT. Note that, the performance of  oracle is our repetition of these methods using corresponding provided code.}
	\label{tab:Lab_ablation}
	\vspace{1mm}
	\setlength{\tabcolsep}{3.0mm}
	\begin{tabular}{l|c|c|c}
		\hline
		\multicolumn{4}{c}{GTA5$\rightarrow$Cityscapes}\\
		\hline
		%Methods & Origin & $+$Trans & Diff. \\
		Methods & Before & $+$Trans & Diff. \\
		\hline
		Direct Transfer & 39.53 & 43.36 & $\uparrow$~3.83\\
		AdaptSeg~\cite{tsai2018learning} & 41.32 & 43.66 & $\uparrow$~2.43 \\
		AdaptSeg-LS~\cite{tsai2018learning} & 43.11 & 45.95 & $\uparrow$~2.84 \\
		Advent~\cite{vu2019advent} & 44.30 & 45.96 & $\uparrow$~1.66\\
		\hline
	\end{tabular}
\vspace{-12pt}
\end{table}
\section{Experiments}
\subsection{Datasets}
In this subsection, we briefly introduce the datasets used to validate our adaptation method, \ie, the widely used dataset Cityscapes~\cite{cordts2016cityscapes}, and recent proposed synthetic datasets GTA5~\cite{Richter_2016_ECCV}, SYNTHIA~\cite{Ros_2016_CVPR}, and Synscapes~\cite{wrenninge2018synscapes}.

\textbf{\textit{Cityscapes}}~\cite{cordts2016cityscapes} consists of 5,000 real-world urban traffic scene images with 2048$\times$1024 resolution and dense-pixel annotation. This dataset is split to 2,975 for training, 500 for validation and 1,525 for testing. Cityscapes annotates 33 categories and 19 of them are used for training and evaluation. The training set without ground truth is used for training adaptation models and validation set for evaluation. 

\textbf{\textit{GTA5}}~\cite{Richter_2016_ECCV} includes 24,966 dense annotated images that are synthesized from a game engine with the resolution of 1914$\times$1052. Its ground-truth labels are consistent with Cityscapes~\cite{cordts2016cityscapes}. In a way similar to previous works~\cite{li2019bidirectional, vu2019advent}, common categories between GTA5 and Cityscapes are used in all experiments. 

\textbf{\textit{SYNTHIA}}~\cite{Ros_2016_CVPR} is a large synthetic dataset that consists of photo-realistic frames rendered from a virtual city. In experiments, we use the SYNTHIA-RAND-CITYSCAPES~\cite{Ros_2016_CVPR} set for adaptation. It contains 9,400 images with a resolution of 1280$\times$760 which are annotated into 16 categories. Similar to GTA5, its annotation are also automatically produced and compatible with Cityscapes. Following previous works~\cite{FDA_Yang_2020_CVPR, tsai2018learning, vu2019advent}, we evaluate performance on 16 and 13 common categories between Cityscapes and SYNTHIA when SYNTHIA is used.

\textbf{\textit{Synscapes}}~\cite{wrenninge2018synscapes} is a synthetic dataset that created using photo-realistic rendering techniques. It consists of 25,000 images at 1440$\times$720 resolution with 33 categories dense annotation and only 19 of them are used. Similarly, its annotations are compatible with Cityscapes. The style of Syncscapes is closer to Cityscapes than GTA5 and Synscapes.

%There are 3 public widely-used synthetic labeled datasets for semantic segmentation domain adaptation, \textit{ie.}, SYNTHIA~\cite{Ros_2016_CVPR}, GTA5~\cite{Richter_2016_ECCV} and Synscapes~\cite{wrenninge2018synscapes}. Since SYNTHIA only includes 16 classes of Cityscapes~\cite{cordts2016cityscapes}, which missing 3 categories label may harmful to the Collaborative and pseudo label generating, we only utilize GTA5 and Synscapes to perform multi-source domain adaptation with our proposed approach.

\subsection{Implementation Details}
We implement our proposed approach with PyTorch and conduct experiments by adopting DeepLab-v2~\cite{chen2017deeplab} with ResNet-101~\cite{he2016deep} as backbone. We report the performance of both initialization from ImageNet pretrained model and initialization from source pretrained model, which are respectively denoted as end-to-end and stage-wise training stratgies. Following prior works~\cite{FDA_Yang_2020_CVPR,vu2019advent,tsai2018learning}, all the networks are trained with stochastic gradient descent~(SGD) optimizer. The initial learning rate and momentum are set to $2.5\times e^{-4}$ and 0.9, respectively, and the polynomial decay policy with power of 0.9 is adopted to adjust the learning rate. We set the batch size to 1 for all datasets during training because of memory limitation. The hyper-parameters in collaborative learning on target domain, $\alpha$ and $\tau$ are set to 50\% and 0.9, respectively. It indicates that we keep pixels that prediction probability within the top 50\% or higher than 0.9 as true labels, and reminder are ignored. The weights $\lambda_S^{col}$ and $\lambda_T^{seg}$ for collaborative learning losses are set to 0.5, 0.1 for end-to-end training and 9.5, 0.1 for stage-wise training. Following Advent~\cite{vu2019advent}, we set the number of maximum iterations to 250,000 but early stop at 120,000 iterations. Same as previous works~\cite{FDA_Yang_2020_CVPR,vu2019advent,tsai2018learning,pan2020unsupervised}, the metric mean intersection-over-union~(mIoU) is used to evaluate the performance of our proposed adaptation method.

\begin{table}
	\centering
	\caption{The validity of model selection and the proposed collaborative learning on the GTA5 $+$ Synscapes to Cityscapes. (a)~shows the performance of each single model and the final ensemble, (b)~shows the comparison of proposed collaborative learning between source domains~(Co-Learning-Src) with baseline and MLDG~\cite{zhang2020generalizable}. \textbf{E:} End-to-End, \textbf{S:} Stage-Wise.}
	\label{tab:col}
	\vspace{1mm}
	\small
	\setlength{\tabcolsep}{0.78mm}
	\begin{tabular}{l|cc}
		\multicolumn{3}{c}{(a)}\\
		\hline
		\hline
		Model & E & S \\
		\hline
		$\M_{S_{GTA5}}$ & 56.90 & 57.72 \\
		$\M_{S_{Syns}}$ & 56.65 & 57.81 \\
		\hline
		$\M_{Ensemble}$ & 58.55 & 59.04 \\
		\hline
	\end{tabular}
\hspace{2mm}
	\begin{tabular}{l|c|c}
		\multicolumn{3}{c}{(b)}\\
		\hline
		%\multicolumn{3}{c}{GTA5+Synscapes $\rightarrow$ Cityscapes}\\
		\hline
		Methods  & mIoU & Diff. \\
		\hline
		Data Combination   & 51.56 & -- \\
		MLDG+TN~\cite{zhang2020generalizable}  & 52.73 & $\uparrow$~1.17\\
		\hline
		%\multirow{2}{*}{Collaborative Learning}  & E & 54.85 & $\uparrow$~3.19\\
		% &  & 54.55 & $\uparrow$~2.99\\
		Co-Learning-Src & 55.79 & $\uparrow$~4.23\\
		%&  & 55.31 & $\uparrow$~3.75\\
		\hline
	\end{tabular}
\vspace{-10pt}
\end{table}

%\begin{table}[t]
%	\centering
%	\caption{Ablation studies of proposed methods. Note that, the performances are achieved by end-to-end training strategy for comparison with simple combination of sources.}
%	\label{tab:combine_ablation}
%	\vspace{1mm}
%	%\renewcommand\arraystretch{1.2}
%	\setlength{\tabcolsep}{.8mm}
%	\begin{tabular}{l|ccc|c}
%		\hline
%		\multicolumn{5}{c}{GTA5 + Synscapes $\rightarrow$ Cityscapes} \\ 
%		\hline
%		%&LAB-based & CL between & CL on & \multirow{2}{*}{mIoU} \\
%		%&Trans. & Source Domains & Target Domain \\
%		& Source &LAB-based & Target & \multirow{2}{*}{mIoU} \\
%		&  Domains &Trans. &  Domain \\
%		%& Trans. & CLSD & CLTD & mIoU \\ 
%		%		\hline
%		%		%\hline
%		%		\multirow{2}{*}{GTA5}&  & & & 39.55\\
%		%		 & $\checkmark$  & & & 44.66\\
%		%		\hline
%		%		\multirow{2}{*}{Synscapes} &  & & & 44.57\\
%		%		 & $\checkmark$  & & & 44.97\\
%		\hline
%		\multirow{2}{*}{Simp-Combination}& \checkmark &  & & 51.59 \\
%		& $\checkmark$ & \checkmark & & 54.38 \\
%		\hline
%		\multirow{3}{*}{Co-Learning} &  & $\checkmark$ & $\checkmark$  & 54.03\\
%		& $\checkmark$ & $\checkmark$ & & 56.03 \\
%		& $\checkmark$ & $\checkmark$ & $\checkmark$ & 58.55 \\
%		\hline
%	\end{tabular}
%\end{table}

\subsection{Ablation Study}
%\begin{table}[t]
%	\centering
%	\caption{The performance of a single models trained with our proposed method and the ensemble of all models.}
%	\label{tab:ens}
%	\begin{tabular}{l|cc}
%		\hline
%		\multicolumn{3}{c}{GTA5 $+$ Synscapes $\rightarrow$ Cityscapes}\\
%		\hline
%		 & E & S\\
%		 \hline
%		 Model1~(GTA5)  & 56.90 & 57.72\\
%		 Model2~(Synscapes) & 56.65 & 57.81 \\
%		 \hline
%		 Ensemble & 58.55 & 59.04  \\ 
%		 \hline
%	\end{tabular}
%\end{table}
In this subsection, extensive experiments on adaptation from GTA+Synscapes to Cityscapes are conducted to study the effectiveness of each component in the proposed approach. 

Firstly, LAB-based image translation is applied to the state-of-the-art UDA methods in semantic segmentation, \ie, AdaptSeg~\cite{tsai2018learning} and Advent~\cite{vu2019advent}, to investigate the effectiveness of proposed image translation method.
Table~\ref{tab:Lab_ablation} is the performance comparison of previous UDA methods training on original and translated images.
From the results, we can see that LAB-based image translation can greatly boost the performance of UDA methods for semantic segmentation. 
For example, LAB-based translation advances the performance of direct transfer from 39.38\% to 43.36\%, which improves about 3.83\%. The performance of AdaptSeg and Advent are also significantly boosted, \ie, 2.43\%, 2.84\% and 1.66\% for AdaptSeg, AdaptSeg-LS and Advent respectively. These results demonstrate that the proposed simple image translation method based on LAB color space is effective for UDA in semantic segmentation. It is easy to implement, with only a little additional computation and without any extra hyper parameter. Note that, all results are reproduced with the code provided by the authors.

\begin{table}[t]
	\centering
	\caption{Ablation studies of proposed methods. Note that, the performances are achieved by end-to-end training strategy for comparison with simple combination of sources.
	}
	\label{tab:combine_ablation}
	\vspace{1mm}
	\setlength{\tabcolsep}{1mm}
	%\small
	\begin{tabular}{cccc|c}
		\hline
		\multicolumn{5}{c}{GTA5 + Synscapes $\rightarrow$ Cityscapes} \\ 
		\hline
		%		LAB-based & Data & CL between & CL on & \multirow{2}{*}{mIoU} \\
		%		Trans. & Comb & Src Dom & Tgt Dom \\
		LAB-based & Data & Co-Learning & Co-Learning &\multirow{2}{*}{mIoU} \\
		Trans. & Comb. & between Src. & on Target & \\
		\hline
		& \checkmark &  & & 51.59 \\
		$\checkmark$ & \checkmark & & & 54.38 \\
		\hline
		$\checkmark$ &  &  & $\checkmark$  & 54.03\\
		$\checkmark$ & & $\checkmark$ & & 56.03 \\
		& & $\checkmark$ & $\checkmark$ &  57.27 \\
		$\checkmark$ & & $\checkmark$ & $\checkmark$ & 58.55 \\
		\hline
	\end{tabular}
	\vspace{-10pt}
\end{table}

\begin{table*}[ht]
	\centering
	\caption{The quantitative comparison with the state-of-the-art methods. DT is the abbreviation of direct transfer. G, S and A indicate GTA5, Synscapes and All respectively. Adv, CL, ST and RL indicate Adversarial learning, Curriculum Learning, Self Training and Reconstruction Learning respectively. Ours-E and Ours-S represent end-to-end training and stage-wise training  of our proposed method respectively.}
	\label{tab:sota_comp}
	\vspace{1mm}
	\small
	\renewcommand\arraystretch{1.05}
	\setlength{\tabcolsep}{0.78mm}{
		\begin{tabular}{l|c|c|ccccccccccccccccccc|c}
			\hline
			\hline
			Methods & \rotatebox{90}{Appr.} & \rotatebox{90}{Source} & \rotatebox{90}{road} & \rotatebox{90}{sidewalk} & \rotatebox{90}{building} & \rotatebox{90}{wall} & \rotatebox{90}{fence} & \rotatebox{90}{pole} & \rotatebox{90}{light} & \rotatebox{90}{sign} & \rotatebox{90}{veg} & \rotatebox{90}{terrain} & \rotatebox{90}{sky} & \rotatebox{90}{person} & \rotatebox{90}{rider} & \rotatebox{90}{car} & \rotatebox{90}{truck} & \rotatebox{90}{bus} & \rotatebox{90}{train} & \rotatebox{90}{mbike} & \rotatebox{90}{bike} & \rotatebox{90}{mIoU} \\
			\hline
			DT~\cite{tsai2018learning} & -- & \multirow{5}{*}{S} & 81.8 & 40.6 & 76.1 & 23.3 & 16.8 & 36.9 & 36.8 & 40.1 & 83.0 & 34.8 & 84.9 & 59.9 & 37.7 & 78.5 & 20.4 & 20.5 & 7.8 & 27.3 & 52.5 & 45.3 \\
			%AdaptSeg~\cite{tsai2018learning} & Adv &  & 94.2 & 60.5 & 85.0 & 29.2 & 25.6 & 39.8 & 43.4 & 43.8 & 85.2 & 35.9 & 88.3 & 63.2 & 41.1 & 87.2 & 30.8 & 44.2 & 29.8 & 28.5 & 53.7 & 53.1\\
			AdaptSeg~\cite{tsai2018learning} & Adv &  & 94.2 & 60.9 & 85.1 & 29.1 & 25.2 & 38.6 & 43.9 & 40.8 & 85.2 & 29.7 & 88.2 & 64.4 & 40.6 & 85.8 & 31.5 & 43.0 & 28.3 & 30.5 & 56.7 & 52.7\\
			FDA~\cite{FDA_Yang_2020_CVPR} & ST & & 93.6 & 58.1 & 84.0 & 30.4 & 29.2 & 39.0 & 43.1 & \textbf{51.7} & 85.9 & 28.8 & 86.9 & 64.0 & \textbf{45.7} & 84.7 & 30.4 & 36.5 & 28.5 & 34.4 & \textbf{62.4} & 53.5\\
			Advent~\cite{vu2019advent} & Adv & & 92.2 & 51.3 & 85.0 & 40.8 & 31.2 & 39.0 & 42.5 & 42.5 & 86.5 & 46.1 & 84.8 & 65.2 & 39.0 & 87.0 & 32.6 & 49.0 & 29.5 & 28.6 & 50.0 & 53.8 \\%dev
			UIA~\cite{pan2020unsupervised} & Adv & & 94.0 & 60.0 & 84.9 & 29.5 & 26.2 & 38.5 & 41.6 & 43.7 & 85.3 & 31.7 & 88.2 & 66.3 & 44.7 & 85.7 & 30.7 & 53.0 & 29.5 & 36.5 & 60.2 & 54.2\\
			\thickhline
			{DT~\cite{tsai2018learning}} & -- & \multirow{8}{*}{G}  & 75.8 & 16.8 & 77.2 & 12.5 & 21.0 & 25.5 & 30.1 & 20.1 & 81.3 & 24.6 & 70.3 & 53.8 & 26.4 & 49.9 & 17.2 & 25.9 & 6.5 & 25.3 & 36.0 & 36.6 \\
			%AdaptSeg~\cite{tsai2018learning} & Adv & & 91.4 & 48.4 & 81.2 & 27.4 & 21.2 & 31.2 & 35.3 & 16.1 & 84.1 & 32.5 & 78.2 & 57.7 & 28.2 & 85.9 & 33.8 & 43.5 & 0.2 & 23.9 & 16.9 & 44.1\\
			AdaptSeg~\cite{tsai2018learning} & Adv & & 86.5 & 25.9 & 79.8 & 22.1 & 20.0 & 23.6 & 33.1 & 21.8 & 81.8 & 25.9 & 75.9 & 57.3 & 26.2 & 76.3 & 29.8 & 32.1 & 7.2 & 29.5 & 32.5 & 41.4\\
			Advent~\cite{vu2019advent} & Adv &  & 89.4 & 33.1 & 81.0 & 26.6 & 26.8 & 27.2 & 33.5 & 24.7 & 83.9 & 36.7 & 78.8 & 58.7 & 30.5 & 84.8 & 38.5 & 44.5 & 1.7 & 31.6 & 32.4 & 45.5\\
			UIA~\cite{pan2020unsupervised} & Adv & &90.6 & 36.1 & 82.6 & 29.5 & 21.3 & 27.6 & 31.4 & 23.1 & 85.2 & 39.3 & 80.2 & 59.3 & 29.4 & 86.4 & 33.6 & 53.9 & 0.0 & 32.7 & 37.6 & 46.3\\
			PyCDA~\cite{Lian_2019_ICCV} & CL &  & 90.5 & 36.3 & 84.4 & 32.4 & 28.7 & 34.6 & 36.4 & 31.5 & 86.8 & 37.9 & 78.5 & 62.3 & 21.5 & 85.6 & 27.9 & 34.8 & 18.0 & 22.9 & 49.3 & 47.4\\
			BDL~\cite{li2019bidirectional} & ST &  & 91.0 & 44.7 & 84.2 & 34.6 & 27.6 & 30.2 & 36.0 & 36.0 & 85.0 & 43.6 & 83.0 & 58.6 & 31.6 & 83.3 & 35.3 & 49.7 & 3.3 & 28.8 & 35.6 & 48.5 \\
			FDA~\cite{FDA_Yang_2020_CVPR} & ST &  & 92.5 & 53.3 & 82.4 & 26.5 & 27.6 & 36.4 & 40.6 & 38.9 & 82.3 & 39.8 & 78.0 & 62.6 & 34.4 & 84.9 & 34.1 & 53.1 & 16.9 & 27.7 & 46.4 & 50.5\\
			PIT~\cite{lv2020spygr} & RL &  & 87.5 & 43.4 & 78.8 & 31.2 & 30.2 & 36.3 & 39.9 & 42.0 & 79.2 & 37.1 & 79.3 & 65.4 & 37.5 & 83.2 & 46.0 & 45.6 & 25.7 & 23.5 & 49.9 & 50.6\\
			\thickhline
			Data Comb. & -- & \multirow{7}{*}{A} & 85.1 & 36.9 & 84.1 & 39.0 & 33.3 & 38.7 & 43.1 & 40.2 & 84.8 & 37.1 & 82.4 & 65.2 & 37.8 & 69.4 & 43.4 & 38.8 & 34.6 & 33.2 & 53.1 & 51.6\\
			%AdaptSeg~\cite{tsai2018learning} & Adv & & 92.4 & 55.4 & 84.7 & 39.2 & 26.6 & 39.0 & 42.0 & 32.2 & 86.0 & 43.6 & 83.8 & 65.6 & 38.4 & 84.1 & 33.1 & 55.4 & 49.6 & 27.7 & 48.9 & 54.1\\%317
			AdaptSeg~\cite{tsai2018learning} & Adv & & 89.3 & 47.3 & 83.6 & 40.3 & 27.8 & 39.0 & 44.2 & 42.5 & 86.7 & 45.5 & 84.5 & 63.1 & 38.0 & 79.4 & 34.9 & 48.3 & 42.1 & 30.7 & 52.3 & 53.7 \\%321
			Advent~\cite{vu2019advent} & Adv & & 91.8 & 49.0 & 84.6 & 39.4 & 31.5 & 39.9 & 42.9 & 43.5 & 86.3 & 45.1 & 84.6 & 65.3 & 41.0 & 87.1 & 37.9 & 49.2 & 31.0 & 30.3 & 48.8 & 54.2\\%318
			MDAN~\cite{zhao2018adversarial} & Adv & & 92.4 & 56.1 & 86.8 & 42.7 & 32.9 & 39.3 & 48.0 & 40.3 & 87.2 & 47.2 & \textbf{90.5} & 64.1 & 35.9 & 87.8 & 33.8 & 48.6 & 39.0 & 27.6 & 49.2 & 55.2\\
			MADAN~\cite{zhao2019multi} & Adv &  & 94.1 & 61.0 & 86.4 & 43.3 & 32.1 & 40.6 & 49.0 & 44.4 & 87.3 & 47.7 & 89.4 & 61.7 & 36.3 & 87.5 & 35.5 & 45.8 & 31.0 & 33.5 & 52.1 & 55.7\\
			%\textbf{COMM-E} & -- &  & 94.1 & \textbf{62.0} & 86.4 & 46.4 & 30.7 & 39.7 & 45.3 & 38.7 & 87.2 & 48.3 & 90.0 & 65.0 & 36.6 & 87.3 & 40.2 & 56.8 & 31.8 & 37.2 & 57.5 & 56.9\\
			%\textbf{COMM-E\dag} & -- &  & 94.1 & 60.3 & 86.6 & \textbf{46.6} & 33.0 & 39.6 & 45.1 & 34.1 & 87.3 & \textbf{49.2} & 89.7 & 65.3 & 34.9 & 88.3 & \textbf{48.7} & \textbf{62.5} & 43.3 & 38.2 & \textbf{58.4} & 58.2 \\
			%\textbf{COMM-E\dag} & -- &  & \textbf{94.2} & \textbf{61.8} & 86.7 & \textbf{47.7} & 34.1 & 39.3 & 44.6 & 34.2 & 87.2 & \textbf{49.6} & 89.7 & \textbf{65.6} & 38.1 & 88.2 & \textbf{48.1} & \textbf{63.0} & 41.9 & \textbf{39.2} & 59.2 & 58.6\\
			\textbf{Ours-E} & -- &  & \textbf{94.2} & \textbf{61.8} & 86.7 & \textbf{47.7} & 34.1 & 39.3 & 44.6 & 34.2 & 87.2 & \textbf{49.6} & 89.7 & \textbf{65.6} & 38.1 & \textbf{88.2} & \textbf{48.1} & \textbf{63.0} & 41.9 & \textbf{39.2} & 59.2 & 58.6\\
			%\textbf{COMM-S} & -- &  & \textbf{94.3} & \textbf{62.6} & 86.8 & 44.1 & 33.8 & \textbf{41.6} & \textbf{49.7} & 42.8 & 87.1 & 48.2 & 90.1 & 63.9 & 38.1 & 87.0 & 38.0 & 58.5 & 45.2 & \textbf{38.7} & 55.8 & 58.2\\
			%\textbf{COMM-S} & -- &  & 92.5 & 54.7 & 86.4 & 42.3 & \textbf{36.8} & 41.1 & 48.2 & 42.2 & 87.3 & 43.8 & 89.1 & 62.2 & 40.3 & \textbf{88.6} & 44.8 & 54.2 & 52.6 & 35.0 & 56.3 & 57.8\\
			%\textbf{COMM-S\dag} & -- &  & 93.6 & 59.6 & \textbf{87.1} & 44.9 & \textbf{36.7} & \textbf{42.1} & \textbf{49.9} & 42.5 & \textbf{87.7} & 47.6 & 89.9 & 63.5 & 40.3 & 88.2 & 41.0 & 58.3 & \textbf{53.1} & 37.9 & 57.7 & \textbf{59.0}  \\
			\textbf{Ours-S} & -- &  & 93.6 & 59.6 & \textbf{87.1} & 44.9 & \textbf{36.7} & \textbf{42.1} & \textbf{49.9} & 42.5 & \textbf{87.7} & 47.6 & 89.9 & 63.5 & 40.3 & 88.2 & 41.0 & 58.3 & \textbf{53.1} & 37.9 & 57.7 & \textbf{59.0}  \\
			\hline
	\end{tabular}}
\vspace{-10pt}
\end{table*}

Because our method can be trained in different strategies and there are $N$ models in the framework, we then test the performance of each single model and the final ensemble one with different training strategies. All these results are based on the setting that adapting from GTA5$+$Synscapes to Cityscapes. As shown in Table~\ref{tab:col}~(a), the performance of all models are significantly improved and the ensembled model achieves the best performance. Therefore, we only report the ensembled model's performance later. The collaborative learning between source domains does not access to the unlabeled target data and therefore can be used to address the task of domain generalization.
Table~\ref{tab:col}~(b) shows the performance comparison with the recent multi-source domain generalization method MLDG~\cite{zhang2020generalizable} and a baseline based on simple data combination. We can see that collaborative learning achieves better generalization performance. For example, MLDG only marginally outperforms the simple baseline about 1.17\%, while collaborative learning boosts the performance about 4.23\%. Note that, collaborative learning here is only applied to source domains without image translation for fair comparison.
%Note that, MLDG would compute and update the parameters of batch normalization layers when inference on target domain, while our method just directly infer on target. 

Table~\ref{tab:combine_ablation} shows different contribution of each component to performance of our proposed approach.
As the results shown, training on combination of source domains can improve the performance on target domain to some extent, which achieves 51.59\% and 54.38\% respectively. Collaborative learning between different source domains further boosts the performance to 56.03\%. Collaborative learning on target domain also boosts the performance on target which achieves 54.03\%. This result shows that collaborative learning between source domains brings more improvements than on target domain. Moreover, full version of our approach achieves the best performance, achieving 58.55\% on target domain. Thus, we can conclude that the proposed approach is effective for unsupervised domain adaptation in semantic segmentation.

\subsection{Comparison with SOTA}

In this subsection, our proposed approach is compared to the recent state-of-the-art single-source and multi-source UDA methods on the GTA5$+$Synscapes to Cityscapes, including a baseline of source-only direct transfer~(DT), single-source UDA methods~\cite{tsai2018learning,pan2020unsupervised,vu2019advent,Lian_2019_ICCV,li2019bidirectional,FDA_Yang_2020_CVPR,lv2020spygr}, the multi-source baseline that simple combination of source domains and multi-source UDA methods~\cite{zhao2018adversarial,zhao2019multi}. We further validate the effectiveness of our proposed method by conducting experiments based on adapting different number of source domains to target domain.

\begin{figure}[t]
	\begin{center}
		\includegraphics[width=1\linewidth]{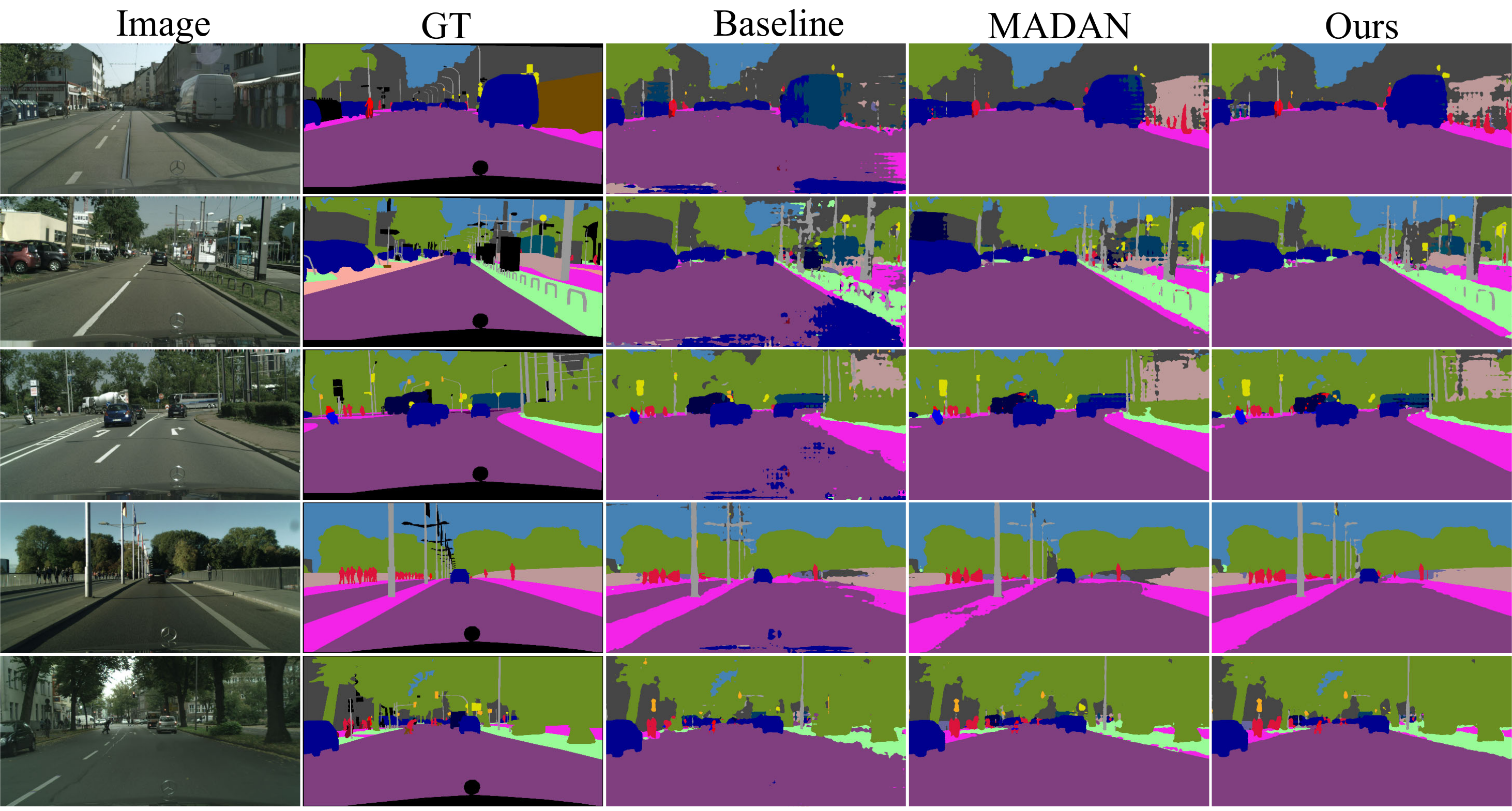}
	\end{center}\vspace{-10pt}
	\caption{Visual Comparison with baseline and other methods. Left to right: Input image from Cityscapes, corresponding ground-truth, segmentation results of baseline that simple combination of source domains, segmentation results of MADAN~\cite{zhao2019multi} and proposed method. Note that, all these results are adapting from GTA5$+$Synscapes.}
	\label{fig:qualitative_comp}
	\vspace{-10pt}
\end{figure}

Table~\ref{tab:sota_comp} shows the results comparison with other methods and Fig.~\ref{fig:qualitative_comp} demonstrates the visual comparison with baseline and MADAN~\cite{zhao2019multi}.
From the results of Table~\ref{tab:sota_comp}, we can see that simply combining the images of source domains for networks training can already greatly boost the generalization performance on target domain, which advances from the 44.1\% of GTA5$\rightarrow$Cityscapes and 45.3\% of Synscapes$\rightarrow$Cityscapes to 51.6\%. Simply adapting UDA methods to train on combination of source domains does not bring much improvements, such as AdaptSeg has only 1\% improvement and Advent only 0.5\%. By employing adaptation strategy, multi-source UDA further boost the performance, \ie, MDAN~\cite{zhao2018adversarial} achieves 55.2\% and MADAN~\cite{zhao2019multi} achieves 55.7\%. By further integrating collaborative learning on the source domains and target domain, we achieve 58.6\% mIoU with the end-to-end training strategy and 59.0\% mIoU with stage-wise training strategy, which
\begin{table}[t]
	\centering
	\caption{The performance of our proposed method that uses different source domains for adaptation. \textbf{G: GTA5, S: Synscapes, Y: SYNTHIA.} mIoU19, mIoU16 and mIoU13 indicate performance on different number of categories.}
	\label{tab:multi-source}
	\vspace{1mm}
	\setlength{\tabcolsep}{1.2mm}
	\begin{tabular}{l|c|ccc}
		\hline
		& sources & mIoU19 & mIoU16 & mIoU13 \\
		\hline
		\multirow{3}{*}{Source-Only} & G & 39.53 & 43.28 & 48.25 \\
		& S & 44.43 & 48.74 & 54.09 \\
		& Y & -- & 32.31 & 37.41 \\
		\hline
		\multirow{4}{*}{Multi-Sources} & G$+$S & 59.04 & 61.25 & 65.87 \\
		& G$+$Y & -- &  54.03 & 59.42\\
		& S$+$Y & -- & 58.19 & 63.18 \\
		& G$+$S$+$Y & -- & 62.24 & 67.15  \\
		\hline
	\end{tabular}
	\vspace{-8pt}
\end{table}
greatly outperforms all the previous methods. When compared to single-source UDA, we observed that our approach achieves more significant improvement on categories such as train, truck, bus and \etal. These objects have rigid body and share much similarity in shape among different source domains. These results also validate the effectiveness of our proposed method. It is noteworthy that our approach does not employ any adversarial learning or any other sophisticated tricks, such as curriculum learning or self-training. More results see supplementary.

Table~\ref{tab:multi-source} shows the performance comparison that adapting different source domains to Cityscapes. As the results shown, our approach can make full use of the labeled source domains and significantly improve the performance on target domain. We can see that adapting from GTA5 and Synscapes achieves 59.04\% mIoU on 19 categories and 61.25\% mIoU on 16 categories, which brings about 15\% and 13\% improvements \textit{w.r.t} to the best model training on single-source. When adapting from all the three labeled source domains, our method further improve the performance~(mIoU on 16 categories) from 61.25\% to 62.24\%. These results further illustrates the effectiveness of our method.
%Note that, because the annotation of SYNTHIA missing 3 categories, we only report 16 and 13 classes of mIoU when source domains include SYNTHIA. 

\section{Conclusion}
In this paper, we present an effective multi-source domain adaptation framework for semantic segmentation based on collaborative learning. A simple image translation method is proposed to reduce the gap between domains. A collaborative learning method based on both labeled source domains and unlabeled target domain is proposed to fully explore essential semantic contexts across domains. Extensive experiments and ablation studies show that the proposed framework is able to significantly outperform all previous state-of-the-arts single-source and multi-source
unsupervised domain adaptation methods, by effectively taking advantage of labeled data from multiple source domains and unlabeled data from target domain.          

{\small
	\bibliographystyle{ieee_fullname}
	\bibliography{egbib}
}

\appendix

\section{More Experiments}
\subsection{Datasets}
\begin{table*}[ht]
	\centering
	\caption{The quantitative results that adapting from GTA5 $+$ Synscapes to IDD and Mapillary respectively. Here, Our-M* means the performance of model $\M_{S_*}$, and Ours-Ensemble means the results that ensemble of all outputs of models $\M_{S_*}$. $\dagger$ means training our proposed approach with stage-wise.}
	\label{tab:sota_comp}
	\vspace{1mm}
	\small
	\renewcommand\arraystretch{1.1}
	\setlength{\tabcolsep}{0.68mm}{
		\begin{tabular}{l|c|c|ccccccccccccccccccc|c}
			\hline
			%\multicolumn{22}{c}{To IDD} \\ 
			\hline
			Methods & \rotatebox{90}{Source} & \rotatebox{90}{Target} & \rotatebox{90}{road} & \rotatebox{90}{sidewalk} & \rotatebox{90}{building} & \rotatebox{90}{wall} & \rotatebox{90}{fence} & \rotatebox{90}{pole} & \rotatebox{90}{light} & \rotatebox{90}{sign} & \rotatebox{90}{veg} & \rotatebox{90}{terrain} & \rotatebox{90}{sky} & \rotatebox{90}{person} & \rotatebox{90}{rider} & \rotatebox{90}{car} & \rotatebox{90}{truck} & \rotatebox{90}{bus} & \rotatebox{90}{train} & \rotatebox{90}{mbike} & \rotatebox{90}{bike} & \rotatebox{90}{mIoU} \\
			\hline
			DT & \multirow{3}{*}{S} & \multirow{15}{*}{\rotatebox{90}{IDD}} & 80.5 & 7.8 & 51.1 & 17.8 & 6.4 & 23.4 & 4.0 & 22.4 & 77.5 & 9.2 & 90.4 & 41.4 & 37.3 & 68.6 & 32.0 & 27.9 & 0.0 & 55.7 & 18.6 & 35.37\\
			AdaptSeg~\cite{tsai2018learning} &  & & 92.5 & 19.4 & 58.1 & 23.2 & 8.9 & 20.4 & 5.0 & 25.7 & 77.2 & 9.5 & 93.9 & 49.6 & 42.7 & 72.0 & 37.1 & 30.6 & 0.0 & 59.6 & 20.0 & 39.23\\
			Advent~\cite{vu2019advent} & & & 93.2 & 19.5 & 59.1 & 21.9 & 8.4 & 23.9 & 5.6 & 24.8 & 79.1 & 9.4 & 94.7 & 48.2 & 40.2 & 71.4 & 37.1 & 29.7 & 0.0 & 58.9 & 21.3 & 39.28\\
			\cline{1-2}\cline{4-23}
			DT & \multirow{3}{*}{G} & & 90.2 & 27.9 & 56.3 & 23.4 & 20.4 & 27.8 & 4.9 & 26.0 & 74.4 & 29.6 & 87.8 & 46.4 & 39.1 & 65.1 & 47.3 & 36.6 & 0.0 & 49.1 & 26.9 & 41.01\\
			AdaptSeg~\cite{tsai2018learning} &  & & 92.8 & 21.4 & 64.7 & 25.0 & 23.3 & 26.9 & 6.0 & 40.7 & 76.7 & 30.5 & 92.5 & 45.7 & 34.0 & 70.9 & 50.5 & 37.5 & 0.0 & 47.6 & 26.2 & 42.78\\
			Advent~\cite{vu2019advent} & & & 93.0 & 25.1 & 66.2 & 31.9 & 22.3 & 29.1 & 10.0 & 38.1 & 73.7 & 26.4 & 93.2 & 49.4 & 43.2 & 72.1 & 52.5 & 40.0 & 0.0 & 50.7 & 26.6 & 44.40\\
			\cline{1-2}\cline{4-23}
			DT & \multirow{9}{*}{S+G} & & 92.2 & 19.1 & 66.0 & 32.1 & 19.4 & 29.4 & 9.5 & 45.1 & 80.3 & 35.7 & 94.8 & 59.4 & 40.5 & 76.4 & 49.3 & 46.6 & 0.0 & 59.9 & 38.4 & 47.06\\
			AdaptSeg~\cite{tsai2018learning} &  & & 92.0 & 18.9 & 66.2 & 23.9 & 17.6 & 30.6 & 5.8 & 45.8 & 81.7 & 30.1 & 94.4 & 57.3 & 47.5 & 75.2 & 51.5 & 53.6 & 0.0 & 58.9 & 35.4 & 46.65\\
			Advent~\cite{vu2019advent} & & & 93.9 & 28.8 & 68.2 & 32.1 & 20.0 & 32.1 & 8.8 & 44.9 & 77.1 & 23.1 & 95.0 & 58.8 & 47.1 & 74.3 & 57.4 & 49.4 & 0.0 & 61.0 & 32.8 & 47.61\\
			\cline{1-1}\cline{4-23}
			Ours-M1 &  & & 95.4 & 38.5 & 70.0 & 36.7 & 21.2 & 25.0 & 14.2 & 43.9 & 78.6 & 28.5 & 94.8 & 58.9 & 45.0 & 70.8 & 56.1 & 48.3 & 0.0 & 63.4 & 38.8 & 48.86 \\
			Ours-M2 & & & 95.1 & 35.2 & 71.2 & 39.0 & 19.3 & 27.2 & 11.5 & 48.1 & 77.8 & 26.3 & 95.3 & 57.6 & 39.2 & 69.7 & 52.2 & 46.1 & 0.0 & 60.0 & 34.0 & 47.63\\
			\textbf{Ours-Ensemble} & & & 95.8 & 41.8 & 72.9 & 39.5 & 21.5 & 26.4 & 18.2 & 44.5 & 78.1 & 28.1 & 95.5 & 62.2 & 43.0 & 70.6 & 58.9 & 49.5 & 0.0 & 63.5 & 38.9 & 49.94\\
			\cline{1-1}\cline{4-23}
			Ours-M1$\dagger$ & & & 95.6 & 39.6 & 71.5 & 38.4 & 19.9 & 30.1 & 12.8 & 47.8 & 78.3 & 31.5 & 95.3 & 55.6 & 47.5 & 74.6 & 48.9 & 54.9 & 0.0 & 64.5 & 39.9 & 49.83\\
			Ours-M2$\dagger$ & & & 95.3 & 37.5 & 71.5 & 36.4 & 21.1 & 31.2 & 13.1 & 44.6 & 79.4 & 33.0 & 95.2 & 55.4 & 46.9 & 73.4 & 51.6 & 44.8 & 0.0 & 64.8 & 41.5 & 49.30\\
			\textbf{Ours-Ensemble}$\dagger$ & & & 95.8 & 39.9 & 73.1 & 38.8 & 21.0 & 31.0 & 14.1 & 43.8 & 78.2 & 32.2 & 95.5 & 58.2 & 47.2 & 74.2 & 52.6 & 50.7  & 0.0 & 65.8 & 41.4 & 50.19\\
			\hline
			%\hline
			%\multicolumn{22}{c}{To Mapillary} \\ 
			\hline
			DT & \multirow{3}{*}{S} & \multirow{15}{*}{\rotatebox{90}{Mapillary}} & 70.4 & 23.6 & 63.6 & 14.8 & 12.0 & 25.8 & 30.7 & 32.7 & 75.2 & 41.2 & 89.4 & 36.2 & 22.0 & 73.0 & 19.5 & 17.2 & 0.2 & 27.7 & 31.1 & 37.18\\
			AdaptSeg~\cite{tsai2018learning} &  & & 85.9 & 24.2 & 73.2 & 17.7 & 27.4 & 26.4 & 33.0 & 39.0 & 75.4 & 44.6 & 94.3 & 34.7 & 27.8 & 77.4 & 25.8 & 16.5 & 1.2 & 29.9 & 31.2 & 41.35\\
			Advent~\cite{vu2019advent} & & & 86.2 & 23.9 & 74.6 & 17.8 & 26.8 & 29.5 & 35.9 & 39.8 & 79.4 & 43.6 & 96.2 & 37.3 & 27.5 & 78.4 & 26.3 & 16.1 & 1.4 & 29.1 & 29.1 & 42.04\\
			\cline{1-2}\cline{4-23}
			DT & \multirow{3}{*}{G} & & 82.2 & 28.6 & 74.2 & 23.4 & 27.2 & 35.3 & 36.4 & 18.6 & 73.8 & 29.2 & 89.6 & 58.9 & 39.2 & 74.5 & 35.0 & 17.2 & 12.5 & 31.3 & 27.8 & 42.89\\
			AdaptSeg~\cite{tsai2018learning} &  & & 86.5 & 31.6 & 78.2 & 24.6 & 30.0 & 36.1 & 35.8 & 31.6 & 73.4 & 33.2 & 93.7 & 59.2 & 44.5 & 78.6 & 41.2 & 39.3 & 14.8 & 36.5 & 32.3 & 47.44\\
			Advent~\cite{vu2019advent} & & & 86.6 & 28.3 & 77.9 & 24.7 & 30.6 & 36.1 & 36.0 & 32.5 & 75.8 & 34.9 & 94.4 & 58.8 & 44.1 & 79.9 & 41.3 & 42.3 & 15.7 & 35.6 & 32.6 & 47.79\\
			\cline{1-2}\cline{4-23}
			DT & \multirow{9}{*}{S+G} & & 77.7 & 30.9 & 75.2 & 27.0 & 27.5 & 33.4 & 37.2 & 37.3 & 76.9 & 43.1 & 93.3 & 55.8 & 38.0 & 72.5 & 38.4 & 40.2 & 2.8 & 36.9 & 42.3 & 46.64\\
			AdaptSeg~\cite{tsai2018learning} &  & & 84.2 & 33.4 & 78.0 & 27.9 & 34.0 & 38.0 & 41.6 & 39.4 & 78.6 & 34.5 & 92.7 & 46.9 & 41.6 & 81.9 & 38.3 & 39.0 & 3.6 & 41.5 & 40.5 & 48.19\\
			Advent~\cite{vu2019advent} & & & 87.2 & 36.2 & 78.0 & 27.1 & 31.2 & 38.4 & 40.8 & 40.2 & 80.8 & 44.2 & 96.0 & 47.1 & 43.5 & 82.3 & 39.0 & 39.3 & 5.0 & 42.0 & 40.3 & 49.40\\
			\cline{1-1}\cline{4-23}
			Ours-M1 & & & 88.2 & 32.5 & 81.0 & 29.1 & 37.5 & 39.9 & 41.7 & 39.6 & 80.4 & 44.6 & 95.8 & 58.7 & 40.2 & 83.1 & 48.1 & 40.7 & 2.3 & 40.1 & 43.2 & 50.89\\
			Ours-M2 & & & 87.8 & 31.6 & 81.0 & 30.0 & 37.8 & 34.8 & 38.3 & 41.3 & 78.1 & 39.1 & 95.1 & 60.1 & 49.5 & 82.2 & 42.7 & 39.0 & 19.2 & 45.9 & 48.0 & 51.67\\
			\textbf{Ours-Ensemble} & & & 88.5 & 34.3 & 81.9 & 31.9 & 41.1 & 39.0 & 40.1 & 41.5 & 79.7 & 45.0 & 95.7 & 62.7 & 51.1 & 83.3 & 49.9 & 45.9 & 8.5 & 46.4 & 47.5 & 53.37 \\
			\cline{1-1}\cline{4-23}
			Ours-M1$\dagger$ & & & 87.5 & 40.1 & 80.9 & 31.0 & 37.4 & 40.0 & 42.5 & 40.6 & 79.6 & 42.4 & 95.2 & 55.5 & 46.5 & 84.5 & 45.1 & 40.3 & 16.5 & 41.6 & 39.1 & 51.92\\
			Ours-M2$\dagger$ & & & 88.6 & 36.5 & 81.4 & 29.7 & 38.2 & 41.3 & 43.0 & 43.4 & 80.2 & 45.8 & 95.6 & 58.3 & 43.8 & 84.5 & 42.5 & 42.0 & 10.1 & 46.2 & 43.9 & 52.37\\
			\textbf{Ours-Ensemble}$\dagger$ & & & 88.4 & 40.1 & 81.9 & 32.4 & 39.8 & 41.4 & 42.2 & 42.7 & 80.1 & 46.4 & 95.6 & 58.2 & 48.5 & 84.7 & 46.6 & 45.5 & 11.7 & 46.9 & 42.4 & 53.44 \\
			\hline
	\end{tabular}}
\end{table*}
Mapillary and IDD are another two widely used benchmarks for autonomous driven scene. They are have more images sampled from more various scenes. Tab.~\ref{tab:data_comp} shows the statistics comparison of different datasets.

\textbf{Mapillary} Vistas dataset~(\textbf{M}) is a large-scale diverse street-level image dataset that containing 25,000 high resolution images with densely pixel-level annotated into 66 object categories. It is designed and compiled to cover diversity, richness of detail and geographic extent. The images are from all around the world, captured at various conditions regarding weather, season and daytime. Moreover, these images come from different imaging devices (mobile phones, tablets, action cameras, professional capturing rigs) and differently experienced photographers. To evaluation our proposed method, we train models with the common 19 categories with Cityscapes~\cite{cordts2016cityscapes} training labels.

\textbf{IDD}~(India Driving Dataset)~\cite{varma2019idd}~(\textbf{I}) consists of 20,000 images, which are obtained from a front facing camera attached to a car and finely annotated with 34 classes collected from 182 drive sequences on Indian roads. Most of images are 1080p resolution with some are 720p. Their label set is expanded in comparison to Cityscapes~\cite{cordts2016cityscapes}, to account for new classes. We train all the models based on the common 19 classes with Cityscapes for adaptation setting. Note that, IDD has another 10k version and here we use thus 20k version one for evalutaion of our proposed method.

\subsection{Results}
Tab.~\ref{Tab:DG_comp} shows the performance comparison of proposed collaborative learning between sources trained on the original images which is not translated with baseline that simple combination and domain generalization method MLDG~\cite{zhang2020generalizable}. From the results, we can see that our proposed collaborative learning can achieve better or comparable performance compared with the state-of-the-art domain generalization method. For example, we achieve 47.80\% and 47.16\% on the IDD and Mapillary dataset, respectively. Both of them are better or comparable to the MLDG.
\begin{table}[t]
	\setlength{\tabcolsep}{0.66mm}
	\caption{The comparison of different datasets for semantic segmentation in autonomous driving.}
	\label{tab:data_comp}
	\begin{tabular}{l|p{1.2cm}<{\centering}p{1.32cm}<{\centering}p{1.4cm}<{\centering}p{1.9cm}<{\centering}}
		\hline
		Dataset & Num. of Images & Num. of Scenes & Cats. (Train/All) & Avg. Resolution \\
		\hline
		Cityscapes~\cite{cordts2016cityscapes} & 5K & 50 & 19/30 & 2048$\times$1024\\
		Mapillary~\cite{MVD2017} & 25K & -- & 19/66 & $\ge$1920$\times$1080\\
		IDD~\cite{varma2019idd} & 20k & 180 & 19/34 & 1678$\times$968\\
		\hline
	\end{tabular}
\end{table}

\begin{table}[t]
	\centering
	\caption{The domain generalization ability comparison of Collaborative Learning Between Sources~(Co-Learning-Src) with baseline and domain generalization method.}
	\label{Tab:DG_comp}
	\setlength{\tabcolsep}{5mm}
	\begin{tabular}{l|c|c}
		\hline
		\multicolumn{3}{c}{GTA5+Synscapes} \\
		\hline
		Method & Target & mIoU \\
		\hline
		Data Combination & \multirow{3}{*}{I} &  47.06\\
		MLDG+TN~\cite{zhang2020generalizable} & & 47.42\\
		Co-Learning-Srcs & & 47.80\\
		\hline
		Data Combination & \multirow{3}{*}{M} & 46.64\\
		MLDG+TN~\cite{zhang2020generalizable} & & 47.11\\
		Co-Learning-Srcs & & 47.16 \\
		\hline
	\end{tabular}
\end{table}

Tab.~\ref{tab:sota_comp} shows the comparison of $\romannumeral1$): the reproduce of AdaptSeg~\cite{tsai2018learning} and Advent~\cite{vu2019advent} that adapting from GTA5, Synscapes and combination of GTA5 and Synscapes to IDD and Mapillary, and $\romannumeral2$): Direct Transfer from GTA5, Synscapes and GTA5$+$Synscapes to IDD and Mapillary, and $\romannumeral3$): each model and ensemble of our proposed method that adapting from GTA5 $+$ Synscapes to IDD and Mapillary. Note that, the network architecture and hyperparameters for different losses are same as the setting to Cityscapes. 

From Tab.~\ref{tab:sota_comp}, we can see that our proposed method achieve the best performance no matter what the target dataset, \ie, achieving 50.19\% and 53.44\% on IDD and Mapillary respectively. Moreover, directly adopting UDA methods on combined sources data sometimes could not achieve better performance than direct transfer. For example, AdaptSeg only achieves 46.65\% when IDD as target domain which is lower the performance of directly transfer based on combined data. All these results further validate the effectiveness of our proposed method.

\end{document}